\documentclass[letterpaper]{article}
\usepackage{uai2018}
\usepackage[margin=1in]{geometry}

\usepackage{times}
\usepackage[utf8]{inputenc} % allow utf-8 input
\usepackage[T1]{fontenc}    % use 8-bit T1 fonts
\usepackage{hyperref}       % hyperlinks
\usepackage{url,color}            % simple URL typesetting
\usepackage{booktabs}       % professional-quality tables
\usepackage{amsfonts}       % blackboard math symbols
\usepackage{nicefrac}       % compact symbols for 1/2, etc.
\usepackage{microtype}      % microtypography
\usepackage{amsmath}
\usepackage{amssymb}
\usepackage{amsthm}
\usepackage{graphicx}
\usepackage{subcaption}
\usepackage{dsfont}
\usepackage{caption}
\usepackage{placeins}
\usepackage{xspace}

\usepackage{natbib}
\setcitestyle{authoryear,open={(},close={)}}

\usepackage{color}
\definecolor{brightmaroon}{rgb}{0.76, 0.13, 0.28}
\definecolor{darkblue}{rgb}{0,0.08,0.45}
\hypersetup{
    colorlinks=true, 
    linkcolor=brightmaroon,  
    citecolor=darkblue,  
    filecolor=brightmaroon,  
    urlcolor=brightmaroon,   
}

\usepackage{algorithm,algorithmic}

\usepackage{enumitem}
\setlist[itemize]{leftmargin=*}

\usepackage[mathcal]{eucal}
\newcommand{\BEAS}{\begin{eqnarray*}}
\newcommand{\EEAS}{\end{eqnarray*}}
\newcommand{\BEA}{\begin{eqnarray}}
\newcommand{\EEA}{\end{eqnarray}}
\newcommand{\BEQ}{\begin{equation}}
\newcommand{\EEQ}{\end{equation}}
\newcommand{\BIT}{\begin{itemize}}
\newcommand{\EIT}{\end{itemize}}
\newcommand{\BNUM}{\begin{enumerate}}
\newcommand{\ENUM}{\end{enumerate}}
\newcommand{\BA}{\begin{array}}
\newcommand{\EA}{\end{array}}
\newcommand{\mysec}[1]{Section~\ref{sec:#1}}

\newcommand{\rb}{\mathbb{R}}

\newcommand{\lossZeroOne}{0\mbox{-}1\xspace}
\usepackage{mathtools}

\newtheorem{proposition}{Proposition}

\title{Marginal Weighted Maximum Log-likelihood for Efficient Learning of Perturb-and-Map models}

\author{ {\bf Tatiana Shpakova} \\
 \\
INRIA - ENS \\ PSL Research University \\
Paris, France \\
\And
{\bf Francis Bach}  \\
\\
INRIA - ENS \\ PSL Research University         \\
Paris, France \\
\And
{\bf Anton Osokin}   \\
\\
National Research University\\
Higher School of Economics\\
Moscow, Russia 
}

\begin{document}

\maketitle

\begin{abstract}
We consider the structured-output prediction problem through probabilistic approaches and generalize the ``perturb-and-MAP'' framework to more challenging weighted Hamming losses, which are crucial in applications. While in principle our approach is a straightforward marginalization, it requires solving many related MAP inference problems. We show that for log-supermodular pairwise models these operations can be performed efficiently using the machinery of dynamic graph cuts. We also propose to use \emph{double} stochastic gradient descent, both on the data and on the perturbations, for efficient learning. Our framework can naturally take weak supervision (e.g., partial labels) into account. We conduct a set of experiments on medium-scale character recognition and image segmentation, showing the benefits of our algorithms.
\end{abstract}

\section{INTRODUCTION}

% Prediction
Structured-output prediction is an important and challenging problem in the field of machine learning. When outputs have a structure, often in terms of parts or elements  (e.g., pixels, sentences or characters), methods that do take it into account typically outperform more naive methods that consider outputs as a set of independent elements. 
Structured-output methods based on optimization can be broadly separated  in two main families: \emph{max-margin  methods}, such as structured support vector machines (SSVM)~\citep{taskar2003max,Tsochantaridis2005} and \emph{probabilistic methods based on maximum likelihoods}  such as conditional random fields (CRF)~\citep{lafferty2001conditional}.

%%Challenges of structured-output
Structured-output prediction faces many challenges: (1)~on top of large input dimensions, problems also have large outputs, leading to scalability issues, in particular when prediction or learning depends on combinatorial optimization problems (which are often polynomial-time, but still slow given they are run many times); (2)~it is often necessary to use losses which go beyond the traditional \lossZeroOne loss to shape the behavior of the learned models towards the final evaluation metric; (3)~having fully labelled data is either rare or expensive and thus, methods should be able to deal with weak supervision.

Max-margin methods can be used with predefined losses, and have been made scalable by several recent contributions~\citep[see, e.g.,][and references therein]{lacosteJulien13BCFW}, but do not deal naturally with weak supervision. 
However, a few works~\citep{yu09learning,kumar10selfpaced,girshick11object} incorporate weak supervision into the max-margin approach via the CCCP~\citep{yuille04concave} algorithm.

% Probabilistic learning
The flexibility of probabilistic modeling naturally allows (a) taking into consideration weak supervision and (b) characterizing the uncertainty of predictions, but it comes with strong computational challenges as well as a non-natural way of dealing with predefined losses beyond the \lossZeroOne loss. 
The main goal of this paper is to provide new tools for structured-output inference with probabilistic models, thus making them more widely applicable, while still being efficient.
There are two main techniques to allow for scalable learning in CRFs: stochastic optimization~\citep{vishwanathan2006accelerated} and piecewise training~\citep{mccallum2005piecewise,sutton2007piecewise,kolesnikov2014closed}; note that the techniques above can also be used  for weak supervision (and we reuse some of them in this work). \

Learning and inference in probabilistic structured-output models recently received a lot of attention from the research community~\citep{Bakir2007book,nowozin2011tutorial,smith2011linguistic}. In this paper 
we consider only models for which maximum-a posteriori (MAP) inference is feasible (a step often referred to as decoding in  max-margin formulations, and which typically makes them tractable). A lot of efforts were spent to explore MAP-solvers algorithms for various problems, leveraging various structures, e.g., graphs of low tree-width~\citep{bishop2006pattern,Wainw08,SontagEtAl_uai08,komodakis2011mrf} and function submodularity~\citep{boros2002pseudo,kolmogorov2004energy,Bach13,osokin2015submodular}. 

Naturally, the existence of even an exact and efficient MAP-solver does not mean that the partition function (a key tool for probabilistic inference as shown below) is tractable to compute. 
Indeed, the partition function computation is known to be $\#P$-hard~\citep{jerrum93polynomial} in general.
For example, MAP-inference is efficient for log-supermodular probabilistic models, while computation of their partition function is not~\citep{djolonga2014MAP}.

For such problems where   MAP-inference is efficient, but partition function computation is not, ``perturb-and-MAP'' ideas such as proposed by~\citet{PaYu11a,hazan12partition} are a very suitable treatment. By adding random perturbations, and then performing  MAP-inference, they can lead to estimates of the partition function. In Section \ref{sec:perturb-and-MAP}, we review the existing approaches to the partition function approximation, parameter learning and inference. 

An attempt to learn parameters via ``perturb-and-MAP'' ideas was made by~\citet{hazan13learning}, where the authors have developed a PAC-Bayesian-flavoured approach for the non-decomposable loss functions. 
While the presented algorithm has something in common with ours (gradient descent optimization of the objective upper bound), it differs in the sense of the objective function and the problem setup, which is more general but that requires a different (potentially with higher variance) estimates of the gradients. Such estimates are usual in reinforcement learning, e.g., the log-derivative trick from the REINFORCE algorithm~\citep{williams1992simple}.

The goal of this paper is to make the ``perturb-and-MAP''  technique applicable to practical problems, in terms of (a) scale, by increasing the problem size significantly, and (b) losses, by treating structured losses such as the Hamming loss or its weighted version, which are crucial to obtaining good performance in practice.

Overall, we make the following contributions:

\vspace*{-.2cm}

\BIT
     \item[--] In Section \ref{sec:marg_lhood}, we generalize the ``perturb-and-MAP'' framework to more challenging weighted Hamming losses which are commonly used in applications. In principle, this is a straightforward marginalization but this requires solving many related MAP inference problems. We show that for graph cuts (our main inference algorithm for image segmentation), this can be done particularly efficiently.
     Besides that, we propose to use a \emph{double} stochastic gradient descent, both on the data and on the perturbations.
    \item[--] In Section \ref{sec:semi_learning}, we show how weak supervision (e.g., partial labels) can be naturally dealt with. Our method in this case relies on approximating marginal probabilities that can be done almost at the cost of the partition function approximation.
    \item[--] In Section \ref{sec:experiments}, we conduct a set of experiments on medium-scale character recognition and image segmentation, showing the benefits of our new algorithms.
\EIT

\section{PERTURB-AND-MAP}
\label{sec:perturb-and-MAP}

In this section, we introduce the notation and review the relevant background.
We study the following probabilistic model (a.k.a.\ a Gibbs distribution) over 
a discrete product space $Y = Y_1 \times \dots \times Y_D$,
\BEQ
P(y) = \frac{1}{Z(f)} e^{f(y)},
\EEQ
which is defined by a potential function $f: Y \to \mathbb{R}$. The constant $Z(f) = \sum\nolimits_{y \in Y}{e^{f(y)}}$ is called the partition function and normalizes $P(y)$ to be a valid probability function, i.e., to sum to one. $Z(f)$ is in general intractable to compute as the direct computation requires summing over exponentially (in $D$) many elements.

Various partition function approximations methods have been used in parameter learning algorithms \citep{parise2005learning}, e.g., mean-field \citep[MF,][]{jordan99bayesian}, tree-reweighted belief propagation~\citep[TRW,][]{Wainw08} or loopy belief propagation~\citep[LBP,][]{Weiss01comparingthe}. 
We will work with the upper bound on the partition function proposed by~\citet{hazan12partition} as it allows us to approximate the partition function via  MAP-inference, calculate gradients efficiently, approximate marginal probabilities and guarantee tightness for some probabilistic models. We introduce this class of techniques below.

\subsection{Gumbel perturbations}
\vspace{-0.1cm}

Recently, \citet{hazan12partition} provided a general-purpose upper bound on the \emph{log-partition function} $A(f) = \log Z(f)$, based on the  ``perturb-and-MAP'' idea~\citep{PaYu11a}: maximize the potential function perturbed by Gumbel-distributed noise.\footnote{The Gumbel distribution on the real line has cumulative distribution function $F(z) = \exp(-\exp(-(z+c)))$, where $c$ is the Euler constant.}  
\begin{proposition}[\citet{hazan12partition}, Corollary 1]
For any function $f: Y \to \rb$, we have $A(f) \leqslant A_{\rm G}(f)$, where
\BEQ
A_{\rm G}(f) = \mathbb{E}_{z_1,\dots,z_D \sim \rm Gumbel} \Big[ \max\limits_{y \in Y} \Big( f(y) +  {\sum_{d=1}^D z_d(y_d)} \Big) \Big].
\EEQ
 {\rm Gumbel} denotes the Gumbel distribution and  $\{z_d(y_d)\}_{y_d \in Y_d}^{d=1,\dots,D}$ is a collection of independent Gumbel samples.
\end{proposition}

\vspace{-0.1cm}

The bound is tight when $f(y)$ is a separable function (i.e., a sum of functions of single variables), and the  tightness of this bound was further studied by \cite{shpakova16parameter} for  log-supermodular models (where $f$ is supermodular). They have shown that the bound $A_G$ is always lower (and thus provide a better bound) than the ``L-field'' bound proposed by \cite{djolonga2014MAP,djolonga2015scalable}, which is itself based on separable optimization on the base polytope of the associated supermodular function.

\vspace{-0.1cm}

The partition function bound $A_G$ can be approximated by replacing the expectation by an empirical average. That is, to approximate it we need to solve a large number (as many as the number of Gumbel samples used to approximate the expectation) of MAP-like problems (i.e., maximizing $f$ plus a separable function) which are feasible by our assumption.
Strictly speaking, the MAP-inference is NP-hard in general, but firstly, it is much easier than the partition function calculation, secondly, there are solvers for special cases, e.g., for log-supermodular models (which include functions $f$ which are negatives of cuts~\citep{kolmogorov2004energy,boykov2004graphcut}) and those solvers are often efficient enough in practice. In this paper, we will focus primarily on a subcase of supermodular potentials, namely negatives of graph cuts.

\vspace{-0.1cm}

\subsection{Parameter learning and Inference}
\label{sec:max_lhood}
\vspace{-0.1cm}

In the standard supervised setting of structured prediction, we are given $N$ pairs of observations $\mathcal{D} = \{ (x^n,y^n)\}_{n=1}^N$, where $x^n$ is a feature representation of the $n$-th object and $y^n \in Y = Y_1 \times \dots \times Y_{D^n}$ is a structured vector of interest (e.g., a sequence of tags, a segmentation MAP or a document summarization representation). 
In the standard linear model, the potential function $f(y| x)$  is represented as a linear combination:  $f(y| x) = w^T\Psi(x,y)$, where $w$ is a vector of  weights and the structured feature map $\Psi(x,y)$ contains the relevant information for the feature-label pair~$(x,y)$. 
To learn the parameters using the predefined probabilistic model, one can use the (regularized) maximum likelihood approach:
\vspace{-0.1cm}
\BEQ
\label{eq:lhood_basic}
\max_w \frac{1}{N} \sum_{n=1}^N \log P(y^n| x^n, w)  - \frac{\lambda}{2} \|w\|^2, 
\EEQ

where $\lambda > 0$ is a regularization parameter
and the likelihood $P(y| x, w)$ is defined as $\frac{\exp( f(y| x) )}{Z(f,x)} = \exp( f(y| x)- A(f,x))$, where $A(f,x)$ is the log-partition function (that now depends on $x$, since we consider conditional models).

\citet{hazan12partition} proposed to learn parameters based on the Gumbel bound $A_G(f, x)$ instead of the intractable log-partition function: 
\begin{align*}
&\log P(y| x) =  f(y| x) - A(f,x) \leq f(y| x) - A_G(f,x) \\[-1mm]
& 
=  f(y| x) - \mathbb{E}_z \Big[\max\limits_{y \in Y}{ \Big\{ \sum_{d=1}^D z_d(y_d) + f(y) \Big\}}  \Big]  \\[-1mm]
& \approx  f(y| x) \!-\! \frac{1}{M} \!\!\sum_{m=1}^M  \max\limits_{y^{(m)} \in Y}{ \Big\{ \sum_{d=1}^D z^{(m)}_d(y^{(m)}_d) + f(y^{(m)})\Big\}}.
\end{align*}

\vspace{-0.1cm}

\citet{hazan12partition} considered the fully-supervised setup where labels $y^n$ were given for all data points $x^n$. \citet{shpakova16parameter} developed the approach, but also considered a setup with missing data (part of the labels $y^n$ are unknown) for the small Weizmann Horse dataset~\citep{borenstein04combining}. 
Leveraging the additional stochasticity present in the Gumbel samples, \citet{shpakova16parameter} extend the use of stochastic gradient descent, not  on the data as usually done, but  on the Gumbel randomization. It is equivalent to the choice of parameter $M = 1$ for every gradient computation (but with a new Gumbel sample at every iteration).
In our work, we use the stochastic gradient descent in a regime stochastic w.r.t.\ \emph{both} the data and the Gumbel perturbations. This allows us to apply the method to large-scale datasets.

\vspace{-0.1cm}

For  linear models, we have $f(y| x) = w^T\Psi(x,y)$ and $\Psi(x,y)$ is usually given or takes zero effort to compute. We assume that the gradient calculation $\nabla_w f(y| x) = \Psi(x,y)$ does not add complexity to the optimization algorithm. 
The gradient of  $\log P(y|x)$ is equal to  $\nabla_w f(y| x) - \nabla_w \max\limits_{y \in Y}{ \Big\{ \sum_{d=1}^D z_d(y_d) + f(y|x)\Big\}} = \nabla_w f(y| x) - \nabla_w f(y^*|x)$, where $y^*$ lies in $\arg \max$ of the perturbed optimization problem.
The gradient of  $\langle \log P(y| x) \rangle $ (the average over a subsample of data, typically a mini-batch) has the form $\langle \nabla_w f(y| x) \rangle - \langle \nabla_w f(y^*| x) \rangle = \langle \Psi(x,y)\rangle_{emp.} - \langle \Psi(x,y^*)\rangle$, where $\langle \Psi(x,y)\rangle_{emp.}$ denotes the empirical average over the data. Algorithm~\ref{algo:sgdGumbel} contains this double stochastic gradient descent (SGD) with stochasticity w.r.t.\ both sampled data and Gumbel samples. The choice of the stepsize $\gamma_h = \frac{1}{\lambda h}$ is standard for strongly-convex problems~\citep{shalev2011pegasos}.

\begin{algorithm}[]
\caption{Double SGD: stochasticity w.r.t.\ data and Gumbel samples\label{algo:sgdGumbel}}
\label{alg:searnn}
\begin{algorithmic}[1]
    \REQUIRE dataset $\mathcal{D} = \{ (x^n,y^n)\}_{n=1}^N$, number of iterations $H$, size of the mini-batch~$T$, stepsize sequence~$\{\gamma_h\}_{h=1}^H$, regularization parameter $\lambda$
    \ENSURE model parameters~$w$
    \STATE \textbf{Initialization}: $w = 0$
    \FOR{$h$ = $1$ \TO $H$}
        \STATE Sample data mini-batch of small size $T$ (that is, $T$ pairs of observations)
        \STATE Calculate sufficient statistics $\langle \Psi(x,y)\rangle_{emp.}$ from the mini-batch
        \FOR{$t$=$1$ \TO $T$}
            \STATE Sample  $z_d(y_d)$ as independent Gumbels for all $y_d \in Y_d$ and for all $d$
            \STATE Find $y^* \in \arg \max_{y \in Y} \big\{ \sum\limits_{d=1}^D z_d(y_d) + f(y) \big\}$
        \ENDFOR
        \STATE Make a gradient step:\\[-7mm]
        $$w_{h+1} \to w_h + \gamma_h \Big{(} \langle \Psi(x,y)\rangle_{emp.} - \langle \Psi(x,y^*)\rangle - \lambda w_h\Big{)} $$
    \vspace{-6mm}\ENDFOR
\end{algorithmic}
\end{algorithm}

\vspace{-0.1cm}

Note, that the classic log-likelihood formulation~\eqref{eq:lhood_basic} is implicitly considering a ``\lossZeroOne loss'' $l_{\lossZeroOne}(y,\hat{y}) = [y \neq \hat{y}] $ as it takes probability of the entire output object~$y^n$ conditioned on the observed feature representation~$x^n$.

\vspace{-0.1cm}

However, in many structured-output problems \lossZeroOne loss evaluation is not an adequate performance measure. The Hamming or weighted Hamming losses that sum mistakes across the $D$ elements of the outputs, are more in demand as they count misclassification per element. 

\vspace{-0.1cm}

\subsection{Marginal probability estimation}
\label{sec:marginals}

\vspace{-0.1cm}

Either at testing time (to provide an estimate of the uncertainty of the model) or at training time (in the case of weak supervision, see \mysec{semi_learning}), we need to  compute marginal probabilities for a single variable $y_d$ out of the $d$ ones, that is,
\vspace{-0.1cm}

$$
P(y_d|  x) = \sum_{ y_{-d}} P(y_{-d}, y_d| x),
$$
where $y_{-d}$ is a sub-vector of $y$ obtained by elimination of the variable $y_d$.  
Following~\cite{hazan12partition} and \citet{shpakova16parameter}, this can be obtained by taking $m$ Gumbel samples and the associated maximizers $y^m \in Y = Y_1 \times \dots \times Y_D$, and, for any particular $d$, counting the number of occurrences in each possible value in all the $d$-th components $y_d^m$ of the maximizers~$y^m$.

\vspace{-0.1cm}

While this provides an estimate of the marginal probability, this is not an easy expression to optimize at it depends on several maximizers of potentially complex optimization problems. In the next section, we show how we can compute a different (and new) approximation which is easily differentiable and on which we can apply stochastic gradient descent. 

\begin{table*}[t]
    \centering
    \caption{
        Variants of the Objective Loss $\ell(w,x,y)$ Function. $\{\theta_d(y_d)\}_{d=1}^{D}$ are the weights of the weighted Hamming loss, $\{q_d(y_d)\}_{d=1}^{D}$ are the marginal probabilities $P(y_d| x)$.}
    \begin{tabular}{|c|c|c|}
        \hline
        Loss & Labelled Data & Unlabelled Data \\ \hline
        \lossZeroOne   & $\log P(w,y| x)$ &  $\log \sum\limits_{y \in Y} P(w,y,x)$  \\ \hline
        Hamming  &$\sum\limits_{d=1}^{D} \log P(w,y_d| x_d)$ & $\sum\limits_{d=1}^{D} \sum\limits_{y_d \in Y_d} q_d(y_d) \log P(w,y_d| x_d)$\\ \hline
        Weighted Hamming   & $\sum\limits_{d=1}^{D} \theta_d(y_d)\log P(w,y_d| x_d)$ & $\sum\limits_{d=1}^{D} \sum\limits_{y_d \in Y_d} q_d(y_d)\theta_d(y_d) \log P(w,y_d| x_d)$ \\ \hline
    \end{tabular}
    \label{tab:obj_func}
\end{table*}

\section{MARGINAL LIKELIHOOD}
\label{sec:marg_lhood}

\vspace{-0.1cm}

In this section, we demonstrate the learning procedure for the element-decoupled losses.
We consider the regularized empirical risk minimization problem in a general form: 
\vspace{-0.1cm}
\BEQ
\label{eq:lhood}
\max_w \frac{1}{N} \sum_{n=1}^N \ell(w, x^n,y^n) - \frac{\lambda}{2} \|w\|^2, 
\EEQ
where $\ell(w,x,y)$ can take various forms from Table~\ref{tab:obj_func} and $\lambda$ is the regularization parameter. 
The choice of the likelihood form is based on the problem setting such as presence of missing data and the considered test-time evaluation function.
\vspace{-0.1cm}
\subsection{Hamming loss}
\label{sec:Hamming}
\vspace{-0.1cm}
The Hamming loss is a loss function that counts misclassification per dimension:
$l_h(y,\hat{y}) = \frac{1}{D}\sum_{d=1}^{D}[y_d \neq \hat{y}_d] $. For this type of loss instead of the classic log-likelihood objective it is more reasonable to consider the following decoupling representation from  Table \ref{tab:obj_func}:
\vspace{-0.1cm}
\BEQ
\textstyle
\ell(w,x,y) = \sum\limits_{d=1}^{D} \log P(y_d| x,w),
\EEQ
\vspace{-0.1cm}
where 
\vspace{-0.1cm}
\BEAS
P(y_d| x,w) & = &\textstyle \sum\limits_{y_{-d}}\!\exp(f(w,y_{-d}| y_d,x)~-~A(f,x)) \\
& = &   \exp(B(f,y_d,x) - A(f,x))
\EEAS
is the marginal probability of the single element $y_d$ given the entire input $x$, 
and $B(f,y_d) = \log \sum_{y_{-d}} \exp(f(w,y_{-d}| y_d)) $, where $y_{-d} \in Y_1 \times \dots Y_{d-1} \times Y_{d+1}  \times\dots \times Y_D$. Thus, the log-marginal probability may be obtained from the difference of two log-partition functions (which we will approximate below with Gumbel samples).
\vspace{-0.1cm}

This idea of considering the marginal likelihood was proposed by \citet{kakade2002alternate}. Our contribution is to consider the approximation by ``perturb-and-MAP'' techniques.
We thus have a new  objective function $\ell(w,x,y)$:
$\ell(w,x,y) = \sum_{d=1}^{D} \left [ (B(f,x,y_d) - A(f,x)) \right ], $
and now the following approximation could be applied:
\vspace{-0.1cm}
\BEAS
A(f) &\!\!\! \approx \!\!\!& A_G(f) = \mathbb{E}_z \Big\{\max_{y \in Y} \sum_{d=1}^D z_d(y_d) + f(y) \Big\}, 
\\[-.1cm]
B(f| y_d) & \!\!\!\approx \!\!\!& B_G(f| y_d) \\[-.1cm]
& \!\!\! = \!\!\!& \mathbb{E}_z \Big\{\max_{y_{-d} \in Y_{-d}} \sum_{s=1: s\neq d}^D z_s(y_s)  + f(y_{-d}| y_d) \Big\}.
\EEAS

\vspace*{-0.25cm}

It is worth noting, that the approximation is not anymore an upper bound of the marginal likelihood; moreover it is a difference of convex functions. 
Remarkably, the objective function exactly matches the log-likelihood in the case of unary potentials (separable potential function) as the log-likelihood function becomes the sum of marginal likelihoods. 
\vspace{-0.1cm}

As noted, the objective $\ell(w,x,y)$ is not convex anymore, but it is presented as the difference of two convex functions.  We can still try to approximate with stochastic gradient descent (which then only converges to a stationary point, typically a local minimum).
Algorithm~\ref{algo:sgdGumbel_Marginal} describes the implementation details.
\vspace{-0.1cm}

\begin{algorithm}[h]
\caption{Double SGD for Marginal Likelihood \label{algo:sgdGumbel_Marginal}}
\begin{algorithmic}[1]
    \REQUIRE dataset $\mathcal{D} = \{ (x^n,y^n)\}_{n=1}^N$, number of iterations $H$, size of the mini-batch~$T$, stepsize sequence~$\{\gamma_h\}_{h=1}^H$, regularization parameter $\lambda$
    \ENSURE model parameters~$w$
    \STATE \textbf{Initialization}: $w = 0$
    \FOR{$h$ = $1$ \TO $H$}
        \STATE Sample data mini-batch of small size $T$ (that is, $T$ pairs of observations)
        \FOR{$t$=$1$ \TO $T$}
            \STATE Sample  $z_d(y_d)$ as independent Gumbels for all $y_d \in Y_d$ and for all $d$
            \STATE Find $y_A^* \in \arg \max_{y \in Y} \big\{ \sum\limits_{d=1}^D z_d(y_d) + f(y) \big\}$
            \FOR{$d$=1 \TO $D$}
            \STATE   Find $y_B^* \in \arg \max\limits_{y_{-d} \in Y_{-d}} \{ \sum\limits_{s=1:s\neq d}^D z_s(y_s) + f(y_{-d}| y_d) \}$
            \ENDFOR
        \ENDFOR
        \STATE Make a gradient step:\\[-7mm]
        $$w_{h+1} \to w_h + \gamma_h \Big{(}\big{\langle} \langle \Psi(x,y_B^*)\rangle - \Psi(x,y_A^*)  \big\rangle- \lambda w_h\Big{)} $$
    \vspace{-6mm}\ENDFOR
\end{algorithmic}
\end{algorithm}
\vspace{-0.1cm}
\paragraph{Acceleration trick.}
Interestingly we can use the same Gumbel perturbation realizations for approximating $A_G(f)$ and $B_G(f| y_d)$ through an empirical average. On the one hand, this restriction should not influence on the result as with a sufficient large averaging number $M$, $A_G(f)$ and $B_G(f| y_d)$ converges to their expectations. This is the same for stochastic gradients: on every iteration, we use a different Gumbel perturbation, but we share this one for the estimation of the gradients of  $A_G(f)$ and $B_G(f| y_d)$. This allows us to save some computations as shown below, while preserving convergence (the extra correlation added by using the same samples for the two gradients  does not change the unbiasedness of our estimates).  
\vspace{-0.1cm}

Moreover, if $y^*_A$ has the same label value $y_d$ as the ground truth, then the MAP inference problem for $y^*_A$ exactly matches the one for $y^*_B$ (with the element $y_d$ fixed from  the ground truth). Then $y^*_A = y^*_B$ and the corresponding difference of gradients gives zero impact into the gradient step. This fact allows us to reduce the number of MAP-inference problems. We should thus calculate $y^*_B$ only for those indices $d$ that leads to a  mismatch between $d$-th label of $y^*_A$ and the ground truth one. 
Remarkably, during the convergence to the optimal value, the reduction will occur more often and decrease the execution time with the number of iteration increase. 
Besides that, in the experiments with graph cuts in Section \ref{sec:experiments} we use dynamic graph cut algorithm for solving several optimization problems of similar structure (here $D$ marginal probabilities calculation). We describe it in more details in Section~\ref{scalable_cuts}.
\vspace{-0.2cm}
\subsection{Weighted Hamming loss}
\label{sec:w_Hamming}
\vspace{-0.1cm}
The weighted Hamming loss is used for performance evaluation in the models, where each dimension has its own level of importance, e.g., in an image segmentation with superpixels, proportional to the size of superpixels. 
It differs from the usual Hamming loss in this way: $l_h(y,\hat{y}) = \frac{1}{D}\sum_{d=1}^{D}\theta_d(y_d)[y_d \neq \hat{y}_d] $. 
\vspace{-0.1cm}

Thus we consider a dimension-weighted model as it can be adjusted for the problem of interest that gives the model more flexibility. 
The optimization problem of interest is transformed from the previous case 
by weighted multiplication: \\[-0.7cm]
\BEQ
\label{eq:w_obj}
\ell(w,x,y) = \sum_{d=1}^{D} \theta_d(y_d) \left [ (B(f,x,y_d) - A(f,x)) \right ].
\EEQ \\[-0.3cm]
To justify this objective function, 
we notice that in the case of unit weights, the weighted loss and objective function~(\ref{eq:w_obj}) match the loss and the objective from the previous section. 
Furthermore, $y_d$ with a large weight $\theta_d(y_d)$ puts more  importance towards  making the right prediction for this $y_d$, and that is why we put more weight on the $d$-th marginal likelihood. This corresponds to the usual rebalancing used in binary classification~\citep[see, e.g.,][and references therein]{bach2006considering}.
Then, the algorithm for this case duplicated the one for the usual Hamming loss and the acceleration trick can be used as well.  
\vspace{-0.1cm}

\subsection{Scalable algorithms for graph cuts}
\label{scalable_cuts}
\vspace{-0.1cm}
As a classical  efficient MAP-solver for pairwise potentials problem we will use graph cut algorithms from~\citet{boykov2004graphcut}. 
The function $f(y| x)$ should then be supermodular, i.e., with pairwise potentials, all pairwise weights of $w$ should remain negative.
\vspace{-0.1cm}

In both Sections \ref{sec:Hamming} and \ref{sec:w_Hamming} we can apply the dynamic graphcut algorithm proposed by~\citet{kohli2007dynamic}, which is a modification of the Boykov-Kolmogorov graphcut algorithms. 
It is dedicated to situations when a sequence of graphcut problems with slightly different unary potentials need to be solved. 
Then, instead of solving them separately, we can use the dynamic procedure and find the solutions for slightly different problems with less costs. 
This makes graphcut scalable for a special class of problems. 
\vspace{-0.1cm}

It can easily be seen that our sequence of problems $y_B^* \in \arg \max\limits_{y_{-d} \in Y_{-d}} \{ \sum\limits_{s=1:s\neq d} z_s(y_s) + f(y_{-d}| y_d) \}$ for $d~=~1,\dots, D$ is a perfect application for the dynamic graph cut algorithm.
At  each iteration we solve $T$ sets of graph cut problems, each of set contains $1+a_t$ problems solvable by the same dynamic cut, where $a_t$ is the number of not matched pixels between $y^*_A$ and ground truth $y^n$.
Finally, using acceleration trick and dynamic cuts we reduce the gradient descent iteration complexity from $\sum_{t=1}^T{(1+D_t)}$ graphcut problems to $T$ dynamic graph cut problems. We make the approach scalable and can apply it for large datasets.
\vspace{-0.1cm}

\section{PARAMETER LEARNING IN THE SEMISUPERVISED SETUP}
\label{sec:semi_learning}
\vspace{-0.1cm}

In this section we assume the presence of objects with unknown labels in the train dataset. We can separate the given data in two parts: fully annotated data $\mathcal{D}_1 = \{ (x^n,y^n)\}_{n=1}^N$ as in the supervised case and unlabeled data $\mathcal{D}_2 = \{x^l\}_{l=1}^L$. Then, the optimal model parameter $w$ is a solution of the following optimization problem: \\[-0.3cm]
\BEQ
\max_w L_1(w) + \kappa L_2(w) - \frac{\lambda}{2}\|w\|^2, 
\EEQ

where $L_1(w) = \sum_{n=1}^N \ell_1(w,x^n,y^n) $, $L_2(w) = \sum_{l=1}^L \ell_2(w,x^l) $ and the parameter $\kappa$ governs the importance of the unlabeled data.  $\ell_1(w,x^n,y^n) $ can have a form from the left column of the Table \ref{tab:obj_func}, and $\ell_2(w,x^l)$ from the right one.
\vspace{-0.3cm}

\paragraph{Marginal calculations.} It is worth   reminding from Section \ref{sec:marginals}, that we can approximate marginal probabilities $q(y)$ of holding $y_d = k$ along with the partition function approximation almost for free.
This can be obtained by taking $m$ Gumbel samples and the associated maximizers $y^m \in Y = Y_1 \times \dots \times Y_D$, and, for any particular $d$, counting the number of occurrences in each possible value in all the $d$-th components $y_d^m$ of the maximizers~$y^m$.
The approximation accuracy depends on number of samples $M$. 
To calculate this we already need to have a trained weight vector $w$ which we can obtained from the fully annotated dataset $\mathcal{D_1} = \{(x^n,y^n\}_{n=1}^N)$. 
We will calculate $q(y)$ for the unlabelled data $\mathcal{D}_2 = \{x^l\}_{l=1}^l$.
Those marginal probabilities contain much more information than MAP inference for the new data as can be seen on the example in Figure \ref{fig:marg_and_MAP_inf_example}. 
We believe that proper use of the marginal probabilities will help to gain better result than using labels from the MAP inference (which we observe in experiments).

It is worth noting that for the inference and learning phases
we use a different number of Gumbel samples.
During the learning phase, we incorporate the double stochastic procedure and use 1 sample per 1 iteration and 1 label. For
the marginal calculation (inference) we should use large
number of samples (e.g. 100 samples) to get accurate approximation.

\vspace{-0.1cm}

\begin{algorithm}[h]
\caption{Sketch for the semisupervised algorithm. \label{algo:semisupervised}}
\begin{algorithmic}[1]
    \REQUIRE fully annotated dataset $\mathcal{D}_1 = \{ (x^n,y^n)\}_{n=1}^N$, number of iterations $H$, size of the mini-batch~$T$, stepsize sequence~$\{\gamma_h\}_{h=1}^H$, regularization param. $\lambda$
    \ENSURE model parameters~$w_1$
    \STATE \textbf{Initialization}: $w_1 = 0$
    \STATE \textbf{Find $w_1$ via Algorithm \ref{algo:sgdGumbel_Marginal}}
    
    \REQUIRE fully annotated dataset $\mathcal{D}_1 = \{ (x^n,y^n)\}_{n=1}^N$, unlabeled dataset $\mathcal{D}_2 = \{x^l\}_{l=1}^L$, number of iterations $H$, size of the mini-batch~$T$, stepsize sequence~$\{\gamma_h\}_{h=1}^H$, regularization parameter $\lambda$
    \ENSURE model parameters~$w_{1,2}$
    \STATE \textbf{Initialization}: $w_{1,2} = w_1$
    \STATE \textbf{Calculate}: $q(y)$ for unlabeled data via $w_1$
    \STATE \textbf{Find $w_{1,2}$ via mixture of Algorithms \ref{algo:sgdGumbel_Marginal} and \ref{algo:unsup}}
\end{algorithmic}
\end{algorithm}
\vspace{-0.1cm}

We provide the sketch of the proposed optimization algorithm in Algorithm~\ref{algo:semisupervised}.
The optimization of $L_1$ is fully supervised and this can be done with tools of the previous section. The optimization of $L_2$ requires the specification of $\ell_2(w,x) $, which we take as $\ell_2(w,x) =\sum\limits_{d=1}^{D} \sum\limits_{y_d \in \{ 0,\dots,K\}} q_d(y_d) \log P(w,y_d| x_d)= \sum\limits_{d=1}^{D} \sum\limits_{y_d \in \{ 0,\dots,K\}} q_d(y_d)B(f| y_d) - DA(f) $, that is, the average of the fully supervised cost function with labels generated from the model $q$. 
The term $L_2$ corresponds to the common way of treating 
unlabeled data through the marginal likelihood.
The sub-algorithm for the optimization of $\ell_2$ is presented as Algorithm \ref{algo:unsup}.
\vspace{-0.1cm}

\begin{algorithm}[h]
\caption{Double SGD for Unsupervised Learning \label{algo:unsup}}

\begin{algorithmic}[1]
    \REQUIRE unlabeled dataset $\mathcal{D}_2 = \{x^l\}_{l=1}^L$, parameter estimate $w_1$, number of iterations $H$, size of the mini-batch~$T$, stepsize sequence~$\{\gamma_h\}_{h=1}^H$, regularization parameter $\lambda$
    \ENSURE model parameters~$w_{1,2}$
    \STATE \textbf{Initialization}: $w_{1,2} = w_1$
    \FOR{$h$ = $1$ \TO $H$}
        \STATE Sample data mini-batch of small size $T$ (that is, $T$ pairs of observations)
        \FOR{$t$=$1$ \TO $T$}
            \STATE Sample  $z_d(y_d)$ as independent Gumbels for all $y_d \in Y_d$ and for all $d$
            \STATE Find $y_A^* \in \arg \max_{y \in Y} \big\{ \sum\limits_{d=1}^D z_d(y_d) + f(y) \big\}$
            \FOR{$d$=1 \TO $D$ \AND $k$=0 \TO $K$}
            \STATE   Find $y_{B,d,k}^* \in \arg \max\limits_{y_{-d} \in Y_{-d}} \{ \sum\limits_{s=1:s\neq d}^{D} z_s(y_s) + f(y_{-d}| y_d=k) \} $
            \ENDFOR
        \ENDFOR
        \STATE Make a gradient step:\\
        $w_{h+1} \to w_h + \gamma_h \Big{(}\big{\langle} \langle \sum_{k=0}^K q_d(k)\Psi(x,y_{B,d,k}^*)\rangle - \Psi(x,y_A^*) \big\rangle - \lambda w_h\Big{)} $
    \ENDFOR
\end{algorithmic}
\end{algorithm}

\vspace{-0.1cm}
\paragraph{Acceleration trick.}
Suppose, that $y_d$ can take values in the range $\{0,\dots,K\}$.
Again we use the same Gumbel perturbation for estimating $A_G(f)$ and $B_{dkG}(f| y_d~=~k)$ for all $k \in \{0,\dots,K\}$.
The consequence of using the same perturbations is that if the $d$-th label $y_d$ of $y^*_A$ takes value $k$, than the corresponding $d$-th gradient will cancel out with one of the $y^*_{Bk}$. 
Thus, we will calculate only $K$ (instead of $K+1$ labels) structured labels  $y^*_{Bl} (l \neq k)$ and reduce the number of optimization problems to be solved. Dynamic graph cuts are applied here as well. 
\vspace{-0.1cm}

Finally in Table \ref{tab:obj_func} we see the relationships between the proposed objective functions. Firstly, the known labels $y^n$ in the supervised case are equivalent to the binary marginal probabilities $q(y^n) \in \{0,1\}^{D^n}$. Secondly, the unit weights $\theta_d(y_d) = 1$ in the weighted Hamming loss are equivalent to the basic Hamming loss.

\paragraph{Partial labels.}
Another case that we would like to mention is  annotation with  partial labels, e.g., in an image segmentation application, the bounding boxes of the images are given.
Then denote $y^{given}$ as the set of given labels.
In this setup the marginal probabilities become conditional ones $q(y_d| y^{given})$ and to approximate this we need to solve several conditional MAP-inference problems. 
The objective function $\ell_2(w) =\sum\limits_{d} \sum\limits_{y_d \in \{ 0,\dots,K\}} q_d(y_d| y^{given}) \log P(w,y_d| x_d,y^{given})$ remains feasible to optimize. 

\section{EXPERIMENTS}
\label{sec:experiments}
\vspace{-0.1cm}
The experimental evaluation consists of two parts: Section
\ref{sec:ocr} is dedicated to the chain model problem, where we compare the  different algorithms for supervised learning; Section \ref{sec:horseseg} is focused on  evaluating our approach for the pairwise model on a weakly-supervised problem.
\vspace{-0.1cm}
\subsection{OCR dataset}
\label{sec:ocr}
\vspace{-0.1cm}
The given OCR dataset from \cite{taskar2003max} consists of handwritten words which are separated in letters in a chain manner. 
The OCR dataset contains 10 folds of $\sim$~6000 words overall.
The average length of the word is $\sim$~9 characters.
Two traditional setups of these datasets are considered: 
1) ``small'' dataset when one fold is considered as a training data and the rest is for test, 2) ``large'' dataset when 9 folds of 10 compose the train data and the rest is the test data.
We perform cross-validation over both setups and present   results in Table \ref{tab:ocr_results}.
\vspace{-0.1cm}

As the MAP oracle we use the dynamic programming algorithm of \cite{viterbi67error}.
The chain structure also allows us to calculate the partition function and marginal probabilities exactly. Thus,  the CRF approach can be applied. We compare its performance with the structured SVM from  \cite{osokin16gapBCFW}, perturb-and-MAP \citep{hazan12partition} and  the one we propose for marginal perturb-and-MAP (as Hamming loss is used for evaluation). 
\vspace{-0.1cm}

The goal of this experiment is to demonstrate that the~CRF approach with exact marginals shows a slightly worse performance as the proposed one with approximated marginals but correct Hamming loss. 
\vspace{-0.1cm}
\begin{table}[h]
\centering
\caption{OCR Dataset. Performance Comparison.}

\vspace*{-.1cm}

\begin{tabular}{c|c|c}
method & small dataset & large dataset  \\ \hline
CRF               & $19.5 \pm  0.4 $ & $ 13.1 \pm 0.8 $ \\ \hline
S-SVM+BCFW        & $19.5 \pm 0.4 $ & $ 12.0 \pm 1.1 $\\ \hline 
perturb\&MAP      & $19.1\pm 0.3$ & $ 12.5 \pm 1.1$\\ \hline 
marg. perturb\&MAP & $19.1\pm 0.3$ & $ 12.8 \pm 1.2$ \\ \hline
\end{tabular}
\label{tab:ocr_results}
\end{table}

For the OCR dataset, we performed 10-fold cross-validation and the numbers of Table~\ref{tab:ocr_results} correspond to the averaged loss function (Hamming loss) values over the 10 folds.
As we can see from the result in  Table~\ref{tab:ocr_results}, the approximate probabilistic approaches slightly outperforms the  CRF on both datasets. The Gumbel approximation (with or without marginal likelihoods) does lead to a better estimation for the Hamming loss. Note that S-SVM performs better in the case of a larger dataset, which might be explained by stronger effects of model misspecification that hurts probabilistic models more than S-SVM~\citep{Pletscher2011}.

\subsection{HorseSeg dataset}
\label{sec:horseseg}
\vspace{-0.1cm}
The problem of interest is   foreground/background superpixel segmentation.
We consider a training set of images $\{x^n\}_{n = 1 \dots N}$ that contain different numbers of superpixels.
A hard segmentation of the image is expressed by an array $y^n \in \{0,1\}^{D^n}$, where $D^n$ is the number of superpixels for the $n$-th image.
\vspace{-0.1cm}

The HorseSeg dataset was created by  \cite{kolesnikov2014closed} and contains horse images. The ``small'' dataset has images with manually annotated labels and contains 147 images.
The second ``medium'' dataset is partially annotated (only bounding boxes are given) and contains 5974 images.
The remaining ``large'' one has 19317 images with no annotations at all.
A fully annotated hold out dataset was used for the test stage. It consists of 241 images. 
\vspace{-0.1cm}

The graphical model is a pairwise model with loops. We  consider  log-supermodular distribution and thus, the max oracle is available as the graph cut algorithm by \cite{boykov2004graphcut}. Note that CRFs with exact inference cannot be used here.
\vspace{-0.1cm}

Following \cite{kolesnikov2014closed}, for the performance evaluation the weighted Hamming loss is used,  where the weight is governed by the superpixel size and foreground/background ratio in the particular image. 
\vspace{-0.1cm}

That is, 
$l_h(y,\hat{y}) = \frac{1}{D}\sum_{d=1}^{D}\theta_d(y_d)[y_d \neq \hat{y}_d], $
where 
$$\theta_d(y_d) = 
\begin{cases}
\displaystyle \frac{V_d}{2V_{foreground}}, & \textrm{if~~} y_d = 1.\\[0.3cm]
\displaystyle \frac{V_d}{2V_{background}}, & \textrm{if~~} y_d = 0 .
\end{cases}$$
$V_d$ is the size of superpixel $d$, $V_{background}$ and $V_{foreground}$ are the sizes of the background and the foreground respectively. In this way  smaller  object sizes have more penalized mistakes.
\vspace{-0.1cm}

Since we incorporate $\theta(y)$ into the learning process and for its evaluation we need to know the background and foreground sizes of the image, this formulation is only applicable for the supervised case, where $y_d$ is given for all superpixels. 
However, in this dataset we have plenty of images with partial or zero annotation. 
For these set of images $\mathcal{D}_2 = \{x^l\}_{l=1}^L$, we handle approximate marginal probabilities $q^l_d$ associated to the unknown labels. Using them we can approximate the foreground and background volumes:
$V^l_{foreground} \approx \sum_{d=1}^{D^l} q^l_d$ and $V^l_{background} \approx \sum_{d=1}^{D^l} (1 - q^l_d)$.
\vspace{-0.1cm}

\begin{figure}

\vspace*{-.2cm}

\centering
\begin{subfigure}{.3\linewidth}
\centering
    \includegraphics[width=2.57cm]{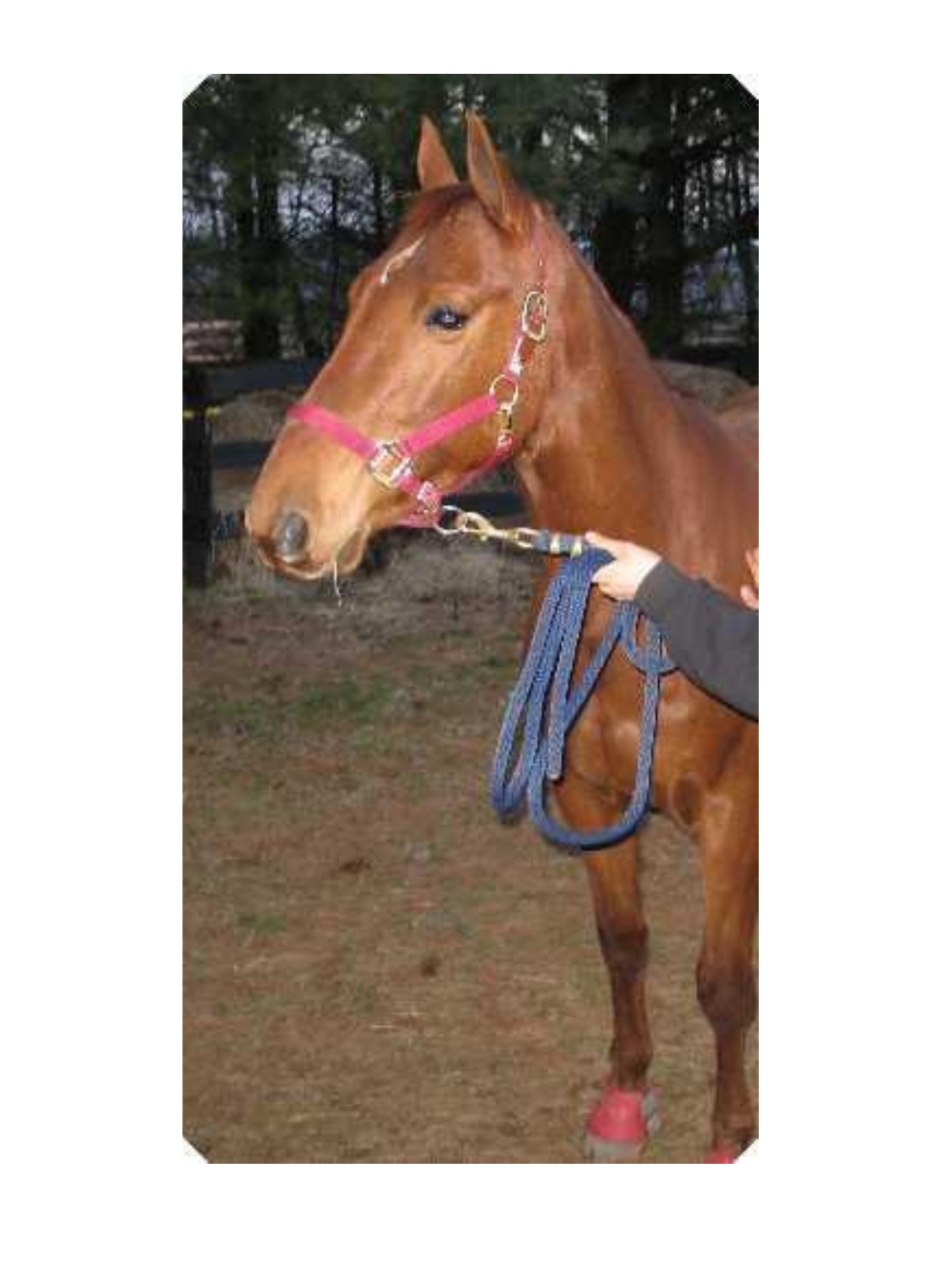}
    \label{fig:original}
\end{subfigure}
\hfill
\begin{subfigure}{.3\linewidth}
\centering
    \includegraphics[width=2.57cm]{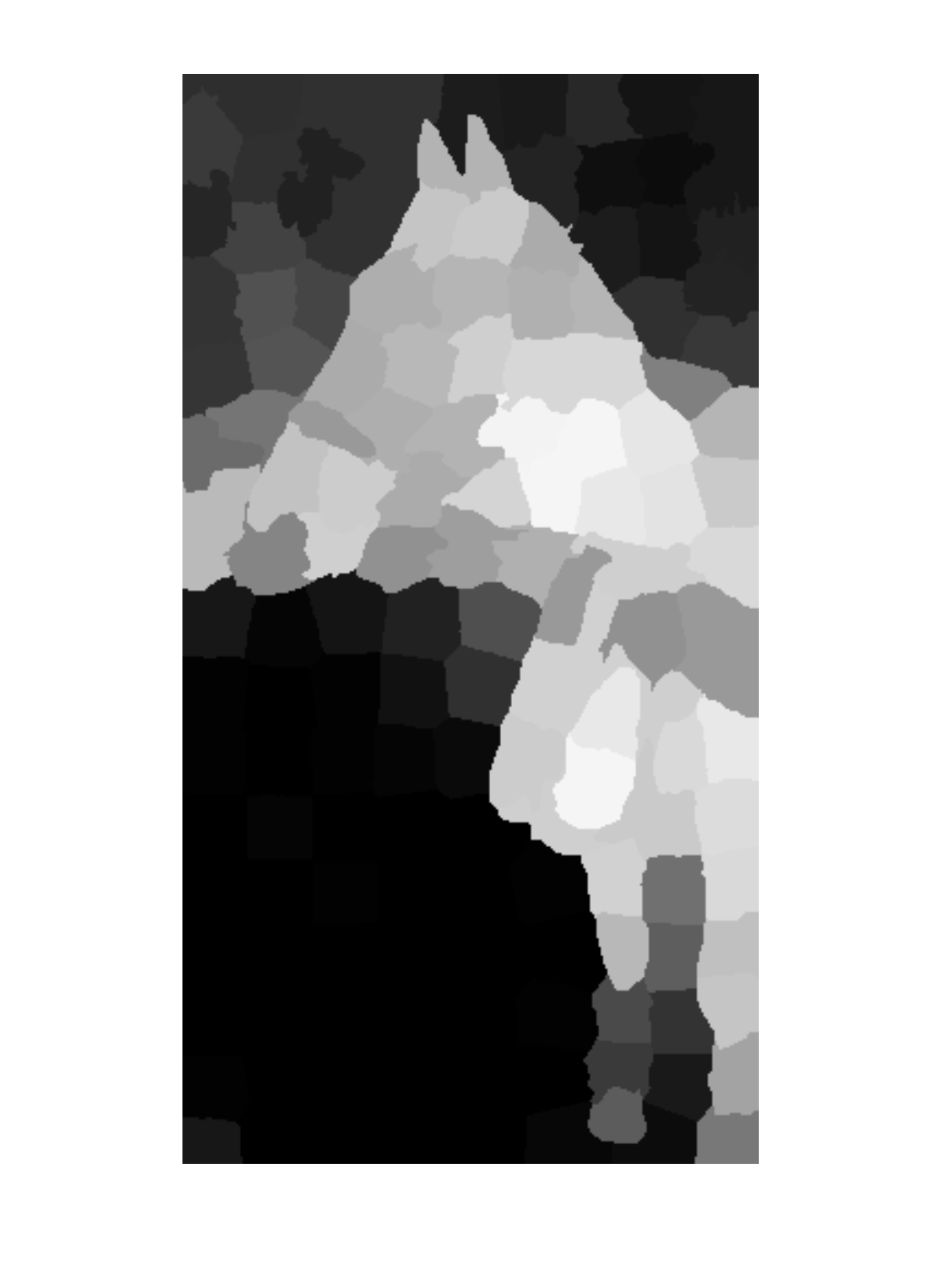}
    \label{fig:marg_inf}
\end{subfigure}
\hfill
\begin{subfigure}{.3\linewidth}
\centering
    \includegraphics[width=2.57cm]{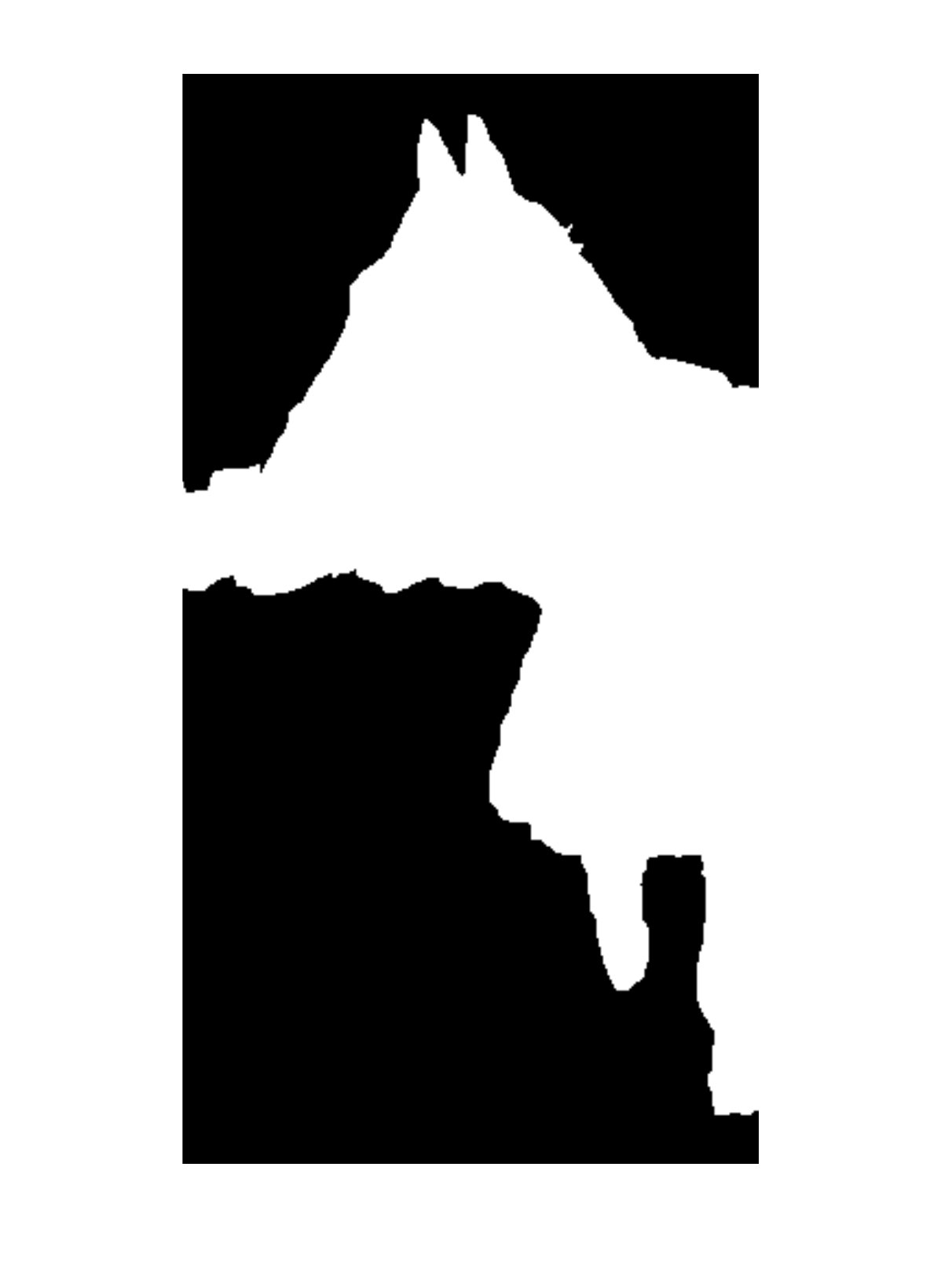}
    \label{fig:MAP_inf}
\end{subfigure}

\vspace*{-.5cm}
\small \ \hspace{-0.7cm} (a) original image \hspace{0.35cm} (b) marginal inf. \hspace{0.5cm} (c) MAP inf.

\caption{ Example of the marginal and MAP inference for an image from the HorseSeg database~\cite{kolesnikov2014closed}.}
\label{fig:marg_and_MAP_inf_example}
\end{figure}

We provide an example of the marginal and MAP inference in Figure \ref{fig:marg_and_MAP_inf_example}.
The difference of the information compression between these two approaches is visually comparable. 
We believe that the   smoother and accurate marginal approach should have a positive impact on the result, as the uncertainty about the prediction is well propagated.
\vspace{-0.1cm}

As an example of the max-margin approaches we take S-SVM+BCFW from the paper of \cite{osokin16gapBCFW} which is well adapted to large-scale problems. 
For S-SVM+BCFW and perturb-and-MAP methods we  use MAP-inference for labelling unlabelled data using $w_1$ (see Section \ref{sec:semi_learning}).
\vspace{-0.1cm}

\normalsize

\begin{table}
    \centering
    \caption{HorseSeg Dataset. Performance Comparison.}
    
    \vspace*{-.1cm}
    
    \begin{tabular}{|c|c|c|c|}
    \hline
         method & ``small'' & ``medium'' & ``large''  \\ \hline
         S-SVM+BCFW & 12.3 & 10.9 & 10.9 \\ \hline
         perturb\&MAP & 20.9 & 21.0 & 20.9 \\ \hline
         w.m. perturb\&MAP & 11.6 & 10.9 & 10.9 \\ \hline 
    \end{tabular}

    \label{tab:horseSeg}
\end{table}

For the HorseSeg dataset (Table~\ref{tab:horseSeg}), the numbers correspond to the averaged loss function (weighted Hamming loss) values over the hold out test dataset.
The results of the experiment in Table \ref{tab:horseSeg} demonstrate that the approaches taking into account the weights of the loss $\theta_d(\cdot)$ (S-SVM+BCFW  and w.m. perturb\&MAP) give a much better accuracy than the regular perturb\&MAP.   
S-SVM+BCFW uses loss-augmented inference and thereby augments the weighted loss structure into the learning phase.
Weighted marginal perturb-and-MAP plugs the weights of the weighted Hamming loss inside the objective log-likelihood function.
Basic perturb-and-MAP does not use the weights $\theta_d(\cdot)$ and loses a lot of accuracy. 
This shows us that the predefined loss for performance evaluation has a significant influence on the result and should be taken into account.
\vspace{-0.3cm}

\paragraph{Small dataset size influence.}
We now investigate the effect of the reduced ``small'' dataset. 
We preserve the setup from the previous section and the only thing that we change is $N$, the size of the ``small'' fully-annotated dataset $\mathcal{D}_1 = \{ (x^n,y^n)\}_{n=1}^N$. The new ``small'' dataset is 10\% the size of the initial one, i.e., only 14 images. By taking a small labelled dataset, we test the limit of supervised learning when few labels are present.

For the HorseSeg dataset (Table~\ref{tab:horseSeg_10percent}), the numbers correspond to the averaged loss function (weighted Hamming loss) values over the hold out test dataset.
The results of this experiments are presented in  Table \ref{tab:horseSeg_10percent}. In this setup, the probabilistic approach ``weighted marginal perturb-and-MAP'' gains more than max-margin ``S-SVM+BCFW''. 
This could happen because of very limited fully supervised data. The learned parameter $w_1$ gives a noisy model and this noisy model produces a lot of noisy labels for the unlabeled data, while weighted perturb-and-MAP is more cautious as it uses probabilities that contain more information (see Figure \ref{fig:marg_and_MAP_inf_example}).

\begin{table}
    \centering
    \caption{Reduced HorseSeg Dataset. Performance Comparison.}
    
    \vspace*{-.1cm}

    \begin{tabular}{|c|c|c|c|}
    \hline
         method & 10\% of  & ``medium'' & ``medium''   \\
          & ``small'' &with bbox &  w/o bbox \\
          \hline
         S-SVM+BCFW & 17.3 & 14.0 & 16.1  \\ \hline
         perturb\&MAP & 23.2 & 23.4 & 23.0 \\ \hline
         w.m. p.\&MAP & 18.1 & 13.7  & 14.4 \\  \hline
    \end{tabular}
    \label{tab:horseSeg_10percent}
\end{table}

\vspace{-0.2cm}

\subsubsection{Acceleration trick impact}
We now compare the execution time of the algorithm
with and without our acceleration techniques (namely
Dynamic Cuts [DC] and Gumbel Reduction[ GR]) to get
an idea on how helpful they are. Table \ref{tab:time_small} shows the execution time for calculating all $y_B^*$ (Algorithm
2) for different numbers of iterations on the HorseSeg small dataset.
We conclude that the  impact of DC does not depend
on the total number of iterations always leading to acceleration around
1.3. For GR, acceleration goes from
3.5 for 100 iterations to 7.6 for one million iterations. Overall,
we get acceleration of factor around 10 for one million
iterations.

\begin{table}[h]
    \centering
    \begin{tabular}{|c|c|c|c|c|c|c|c|}
    \hline
        method $\backslash$ it & 100 & $10^3$ & $10^4$ & $10^5$ & $10^6$  \\ \hline
        basic & 0.9 & 9.2 & 89.5 & 900 & 8993    \\ \hline
        DC & 0.7 & 6.9 & 69.0 & 696 & 7171  \\ \hline
        GR & 0.3 & 2.1 & 15.5 & 133.4 & 1186 \\ \hline
        DC+GR & 0.2 & 1.5 & 10.9 & 83.5 & 727 \\ \hline
    \end{tabular}
    \caption{Execution time comparison in seconds. HorseSeg small dataset.}
    \label{tab:time_small}
\end{table}

\subsection{Experiments analysis}
\vspace{-0.2cm}

The experiments results mainly show that not taking into account the right loss in the learning procedure is detrimental to probabilistic technique such as CRFs, while taking it into account (our novelty) improves results. Also, Tables~\ref{tab:ocr_results} and~\ref{tab:horseSeg} show that the proposed methods achieves (and sometimes surpasses) the level performance of the max-margin approach (with loss-augmented inference). 

Further, we observed that the size of the training set influences the SSVM and perturb-and-MAP approaches differently. For smaller datasets, the max-margin approaches tend to lose information due to usage of the hard estimates for the unlabelled data (e.g. in Table~\ref{tab:horseSeg_10percent}: 16.1 against 14.4 for “medium” dataset without bounding boxes labeling).

Table~\ref{tab:horseSeg_10percent} reports an experiment about using weakly-labeled data at the training stage (the results on the partially annotated ``medium'' dataset). This experiment studied the impact on the final prediction quality of the training set of “medium” size on top of the reduced “small” fully-labelled set. The results of Table~\ref{tab:horseSeg_10percent} mean that the usage of our approach adopted to the correct test measure outperforms the default perturb-and-MAP by a large margin. Our approach also significantly outperformed the comparable baseline of SSVM due to reduced size of the “small” fully-labelled set.

\vspace{-0.1cm}

\section{CONCLUSION}
\vspace{-0.1cm}
In this paper, we have proposed an approximate learning technique for problems with non-trivial losses.
We were able to make marginal weighted log-likelihood for perturb-and-MAP tractable.
Moreover, we used it for semi-supervised and weakly-supervised learning.
Finally, we have successfully demonstrated good performance of the marginal-based and weighted-marginal-based approaches on the middle-scale experiments. 
As a future direction, we can go beyond the graph cuts and image segmentation application and consider other combinatorial problems with feasible MAP-inference, e.g., matching.

\subsubsection*{Acknowledgements}
We acknowledge support the European Union's H2020 Framework Programme (H2020-MSCA-ITN-2014) under grant agreement n$^\text{o}$642685 MacSeNet. 
This research was also partially supported by Samsung Research, Samsung Electronics, and the European Research Council (grant SEQUOIA 724063).

\vspace{-0.2cm}

\bibliography{bib}

\end{document}